\documentclass{article}

\usepackage{microtype}
\usepackage{graphicx}
\usepackage{amsmath}
\usepackage{amsfonts}
\usepackage{graphicx}
\usepackage{caption}
\usepackage{subcaption}
\usepackage{algorithm}
\usepackage{booktabs} 

\usepackage[accepted]{icml2018}

\icmltitlerunning{Active Testing: An Efficient and Robust Framework for
Estimating Accuracy}
\begin{document}

\twocolumn[
\icmltitle{Active Testing: An Efficient and Robust Framework for Estimating
Accuracy.}


\icmlsetsymbol{equal}{*}

\begin{icmlauthorlist}
\icmlauthor{Phuc Nguyen}{uci}
\icmlauthor{Deva Ramanan}{cmu}
\icmlauthor{Charless Fowlkes}{uci}
\end{icmlauthorlist}

\icmlaffiliation{uci}{University of California, Irvine}
\icmlaffiliation{cmu}{Carnegie Mellon University}
\icmlcorrespondingauthor{Phuc Nguyen}{nguyenpx@uci.edu}
\icmlkeywords{active learning, computer vision}

\vskip 0.3in
]


\printAffiliationsAndNotice{}  

\begin{abstract}
    Much recent work on visual recognition aims to scale up
    learning to massive, noisily-annotated datasets. We address the problem of
    scaling-up the {\em evaluation} of such models to large-scale datasets with
    noisy labels.  Current protocols for doing so require a human user to
    either vet (re-annotate) a small fraction of the test set and ignore the
    rest, or else correct errors in annotation as they are found through manual
    inspection of results. In this work, we re-formulate the problem as one of
    {\em active testing}, and examine strategies for efficiently querying a user so
    as to obtain an accurate performance estimate with minimal vetting. We
    demonstrate the effectiveness of our proposed active testing framework on
    estimating two performance metrics, Precision@K and mean Average Precision,
    for two popular computer vision tasks, multi-label classification and
    instance segmentation. We further show that our approach is able to
    save significant human annotation effort and is more robust than
    alternative evaluation protocols.
\end{abstract}

\section{Introduction}
Visual recognition is undergoing a period of transformative progress, due in
large part to the success of deep architectures trained on massive datasets
with supervision. While visual data is in ready supply, high-quality supervised
labels are not. One attractive solution is the exploration of unsupervised
learning.  However, regardless how they are trained, one still needs to {\em
evaluate} accuracy of the resulting systems.  Given the importance of rigorous,
empirical benchmarking, it appears impossible to avoid the costs of assembling
high-quality, human-annotated test data for test evaluation.

\begin{figure}[t!]
    \centering
    \includegraphics[width=\linewidth]{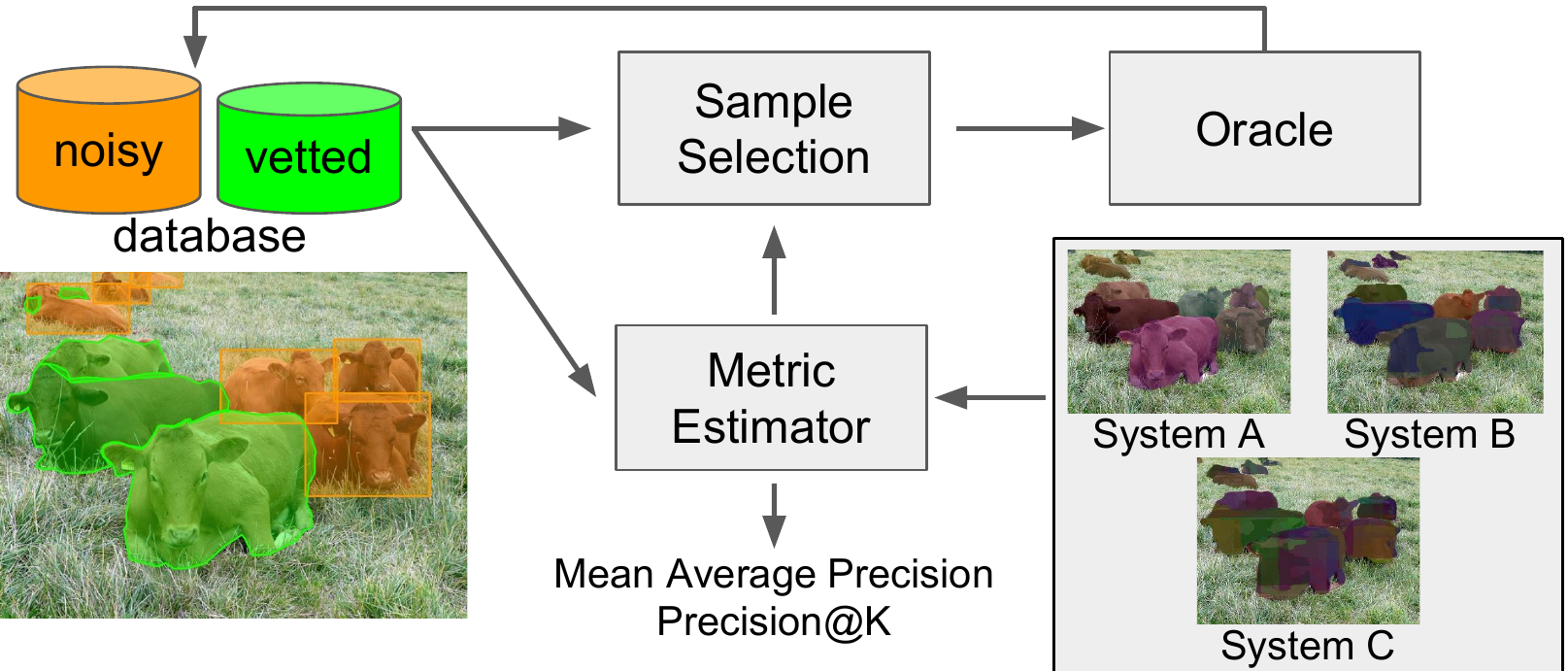}
    \caption{Classic methods for benchmarking algorithm performance require a
    test-set with high-quality labels. While it is often easy to obtain
    large-scale data with noisy labels, test evaluation is typically carried
    out on only a small fraction of the data that has been manually cleaned-up
    (or ``vetted'').  We show that one can obtain dramatically more accurate
    estimates of performance by using the vetted-set to train a statistical
    estimator that both (1) reports improved estimates and (2) actively selects
    the next batch of test data to vet. We demonstrate that such an
    ``active-testing'' process can efficiently benchmark performance and 
    and rank visual recognition algorithms.}
\label{fig:overview}
\vspace{-0.2in}
\end{figure}

Unfortunately, manually annotating ground-truth for large-scale test datasets
is often prohibitively expensive, particularly for rich annotations required to
evaluate object detection and segmentation.  Even simple image tag annotations
pose an incredible cost at scale~\footnote{For example, NUS-WIDE,
\cite{chua2009nus} estimated 3000 man-hours to semi-manually annotate a relatively small set of 81 concepts across 270K
images}.  In contrast, obtaining noisy or partial annotations is often far
cheaper or even free.  For example, numerous social media platforms produce
image and video data that are dynamically annotated with user-provided tags
(Flickr, Vine, Snapchat, Facebook, YouTube).  While much work has explored the
use of such massively-large ``webly-supervised'' data sources for
learning~\cite{wu2015ml, yu2014large,li2017learning,veit2017learning}, we
instead focus on them for evaluation.

How can we exploit such partial or noisy labels during testing?  With a limited
budget for vetting noisy ground-truth labels, one may be tempted to simply
evaluate performance on a small set of clean data, or alternately just trust
the cheap-but-noisy labels on the whole dataset. However, such approaches can
easily give an inaccurate impression of system performance.  We show in our
experiments that these naive approaches can produce alarmingly-incorrect
estimates of comparative model performance. Even with a significant fraction of
vetted data, naive performance estimates can incorrectly rank two algorithms in
15\% of trials, while our active testing approach significantly reduces this misranking error to 3\%.

The problem of label noise even exists for ``expertly'' annotated datasets,
whose construction involves manual selection of a test set which is deemed
representative in combination with crowd-sourced labeling by multiple
experts~\cite{rashtchian2010collecting,khattak2011quality}. Preserving
annotation quality is an area of intense research within the HCI/crowdsourcing
community~\cite{kamar2012combining,sheshadri2013square}. In practice,
annotation errors are often corrected incrementally through multiple rounds of
interactive error discovery and visual inspection of algorithm test results
over the lifetime of the dataset.  For example, in evaluating object
detectors, the careful examination of detector errors on the test
set~\cite{hoiem2012diagnosing} often reveals missing annotations in widely-used
benchmarks~\cite{lin2014microsoft,everingham2015pascal,dollar2012pedestrian}
and may in turn invoke further iterations of manual corrections
(e.g.,~\cite{mathias2014face}).  In this work, we formalize such ad-hoc
practices in a framework we term {\em active testing}, and show that
significantly improved estimates of accuracy can be made through simple
statistical models and active annotation strategies.

\section{Related Work}

\textbf{Benchmarking:}
Empirical benchmarking is now widely considered to be an integral tool in the
development of vision and learning algorithms.  Rigorous evaluation, often in
terms of challenge
competitions~\cite{russakovsky2015imagenet,everingham2010pascal} on held-out
data, serves to formally codify proxies for scientific or application goals and
provides quantitative ways to characterize progress towards them.  The
importance and difficulties of test dataset construction and annotation are now
readily appreciated~\cite{ponce2006dataset,torralba2011unbiased}.

Benchmark evaluation can be framed in terms of the well-known {\em empirical
risk} minimization approach to learning~\cite{vapnik1992principles}.
Benchmarking seeks to estimate the risk, defined as the expected loss of an
algorithm under the true data distribution. Since the true distribution is
unknown, the expected risk is estimated by computing loss a finite sized sample
test set.  Traditional losses (such as 0-1 error) decompose over test examples,
but we are often interested in multivariate ranking-based metrics that do not
decompose (such as Precision@K and Average Precision~\cite{joachims2005support}). Defining and estimating expected risk
for such metrics is more involved (e.g., Precision@K should be replaced by
precision at a specified quantile ~\cite{boyd2012accuracy}) but generalization
bounds are known ~\cite{agarwal2005generalization,hill2002average}.  For
simplicity, we focus on the problem of estimating the empirical risk on a
fixed, large but finite test set.

\begin{figure}[t]
    \centering
    \includegraphics[width=0.9\linewidth]{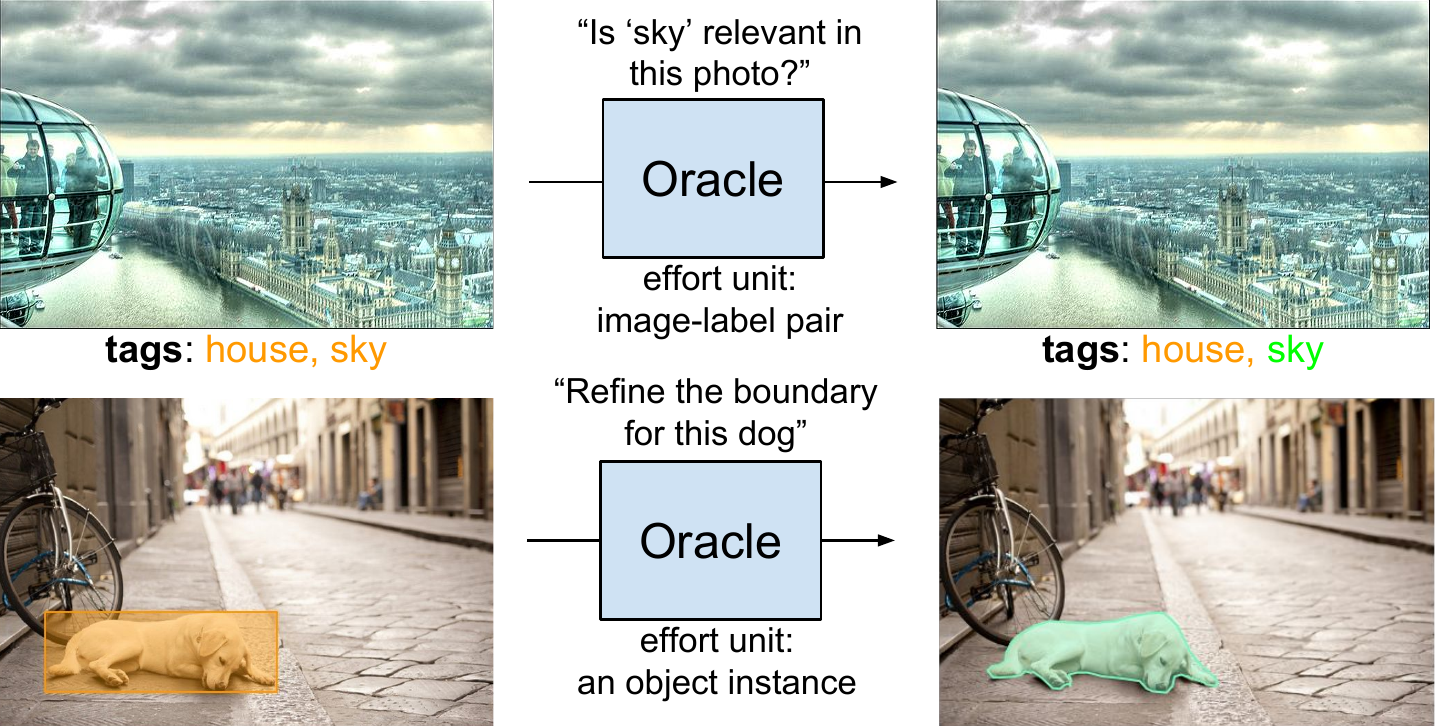}
    \caption{Vetting Procedure. The figure shows the vetting procedures for the
    multi-label classification (top) and instance segmentation (bottom) tasks.
    The annotations on the left are often incomplete and noisy, but
    significantly easier to obtain. These initial noisy annotations are
    ``vetted'' and corrected if necessary by a human.  We quantify human
    effort in units of the number of image-label pairs corrected or object
    segment masks specified.}
\label{fig:vetting_procedures}
\vspace{-0.25in}
\end{figure}

\textbf{Semi-supervised testing:} To our knowledge, there have only been a
handful of works specifically studying the problem of estimating recognition
performance on partially labeled test data.  Anirudh et
al.~\cite{anirudh2014interactively} study the problem of 'test-driving' a
detector to allow the users to get a quick sense of the generalizability of the
system.  Closer to our approach is that of Welinder et
al.~\cite{welinder2013lazy}, who estimate the performance curves using a
generative model for the classifier's confidence scores. Their approach
leverages ideas from the semi-supervised learning literature while our approach
builds on active learning. 

The problem of estimating benchmark performance from sampled relevance labels
has been explored more extensively in the information retrieval literature
where complete annotation was acknowledged as infeasible.  Initial work focused
on deriving labeling strategies that produce low-variance and unbiased
estimates \cite{yilmaz2006estimating,aslam2006statistical} and identifying
performant retrieval systems \cite{moffat2007strategic}.
\cite{sabharwal2017good} give error bounds for estimating PR and ROC curves by
choosing samples to label based on the system output ranking.
\cite{gao2014reducing} estimate performance using an EM algorithm to integrate
relevance judgements.  \cite{li2017active} and \cite{rahman2018efficient} take
a  strategy similar to ours in actively selecting test items to label as well
as estimating performance on remaining unlabeled data.

\textbf{Active learning:} Our proposed formulation of active testing is closely
related to active learning. From a theoretical perspective, active learning can
provide strong guarantees of efficiency under certain
restrictions~\cite{balcan2016active}.  Human-in-the-loop active learning
approaches have been well explored for addressing training data collection in
visual recognition systems~\cite{branson2010visual,wah2011multiclass,vijayanarasimhan2014large}.
One can view {\em active testing} as a form of active learning where the
actively-trained model is a statistical predictor of performance on a test set.
Active learning is typically cast within the standard machine-learning
paradigm, where the goal is to (interactively) learn a model that makes
accurate per-example predictions on held-out {\em i.i.d} data. In this case,
generalization is of paramount importance. On the other hand, active-testing
interactively learns a model that makes {\em aggregate} statistical predictions
over a {\em fixed} dataset. This means that models learned for active-testing
(that say, predict average precision) need not generalize beyond the test set
of interest. This suggests that one can be much more aggressive in overfitting
to the statistics of the data at hand.

\section{Framework for Active Testing}
In this section, we introduce the general framework for active testing.
Figure~\ref{fig:overview} depicts the overall flow of our approach.  Our
evaluation database initially contains test examples with inaccurate (noisy)
annotations. We select a batch of data items whose labels will be manually
vetted by an oracle (e.g., in-house annotators or a crowd-sourced platform such
as Mechanical Turk). Figure~\ref{fig:vetting_procedures} shows examples of such
noisy labels and queries to Oracle. The evaluation database is then updated
with these vetted labels to improve estimates of test performance. Active
testing consists of two key components: a {\em metric estimator} that estimates
model performance from test data with a mix of noisy and vetted labels, and a
{\em vetting strategy} which selects the subset of test data to be labeled in
order to achieve the best possible estimate of the true performance.

\subsection{Performance Metric Estimators}

We first consider active testing for a simple binary prediction problem and
then extend this idea to more complex benchmarking tasks such as multi-label
tag prediction and instance segmentation.  As a running example, assume that we
are evaluating an system that classifies an image (e.g., as containing a cat or
not).  The system returns of confidence score $s_i \in \mathbb{R}$ for each
test example $i \in \{1 \ldots N\}$.  Let $y_i$ denote a ``noisy'' binary label
for example $i$ (specifying if a cat is present), where the noise could arise
from labeling the test set using some weak-but-cheap annotation technique
(e.g., user-provided tags, search engine results, or approximate annotations).
Finally, let $z_i$ be the true latent binary label whose value can be obtained
by rigorous human inspection of the test data item.

Typical benchmark performance metrics can be written as a function of the true
ground-truth labels and system confidences. We focus on metrics that only
depend on the rank ordering of the confidence scores and denote such a metric
generically as $Q(\{z_i\})$ where for simplicity we hide the dependence on $s$
by assuming that the indices are always {\em sorted} according to $s_i$ so that
$s_1 \geq \cdots \geq s_N$.  For example, commonly-used metrics for binary
labeling include precision@K and average precision (AP):
\begin{align}
  Prec@K(\{z_1, \ldots, z_N\}) &= \frac{1}{K}\sum_{i \leq K} z_i \\
  AP(\{z_1, \ldots, z_N\}) &= \frac{1}{N_{p}} \sum_k \frac{z_k}{k} \sum_{i \leq k} z_i
\end{align}
where $N_{p}$ is the number of positives.  We include derivations in
supplmental material.

{\bf Estimation with partially vetted data:}
In practice, not all the data in our test set will be vetted. Let us divide the
test set into two components, the unvetted set $U$ for which we only know the
approximate noisy labels $y_i$ and the vetted set $V$, for which we know the
ground-truth label. With a slight abuse of notation, we henceforth treat the
true label $z_i$ as a random variable, and denote its observed realization (on
the vetted set) as $\tilde{z}_i$.  The simplest strategy for estimating the
true performance is to ignore unvetted data and only measure performance $Q$
on the vetted subset:
\begin{align}
  Q(\{\tilde{z_i}: i \in V \}) \qquad \text{{\bf [Vetted Only]}}
\end{align}
This represents the traditional approach to empirical evaluation in which we
collect a single, vetted test dataset and ignore other available test data.
This has the advantage that it is unbiased and converges to the true empirical
performance as the whole dataset is vetted. The limitation is that it makes
use of only fully-vetted data and the variance in the estimate can be quite
large when the vetting budget is limited.

A natural alternative is to incorporate the unvetted examples by simply
substituting $y_i$ as a ``best guess'' of the true $z_i$. We specify this
{\em naive} assumption in terms of a distribution over all labels $z = \{z_1,\ldots,z_N\}$:
\begin{align}
  p_{naive}(z) = \prod_{i \in U} \delta (z_i = y_i) \prod_{i \in V}  \delta(z_i = \tilde{z}_i)
\end{align}
where $\tilde{z}_i$ is the label assigned during vetting. Under this assumption
we can then compute an expected benchmark performance:
\begin{align}
  E_{p_{naive}(z)} \Big[ Q(z) \Big] \qquad \text{{\bf [Naive Estimator]}}
\end{align}
which amounts to simply substituting $\tilde{z}_i$ for vetted examples and
$y_i$ for unvetted examples.

Unfortunately, the above performance estimate may be greatly affected by noise
in the nosiy labels $y_i$. For example, if there are systematic biases in the
$y_i$, the performance estimate will similarly be biased. We also consider more general scenarios where side information such as features of the test items and distribution of scores of the classifier under test may also be informative.  We thus propose computing the expected performance under a more sophisticated estimator:
\begin{align}
p_{est}(z) = \prod_{i \in U} p(z_i | \mathcal{O})  \prod_{i \in V}  \delta(z_i = \tilde{z}_i) 
\end{align}
where $\mathcal{O}$ is the total set of all observations available to the
benchmark system (e.g. noisy labels, vetted labels, classifier scores, data features).  We make the plausible assumption that the distribution of unvetted labels factors conditioned on $\mathcal{O}$.

\begin{algorithm}[tb]
   \caption{Active Testing Algorithm}
   \label{alg:active_testing}
{\small
\begin{algorithmic}
   \STATE {\bfseries Input:} unvetted set $U$, vetted set $V$, total budget $T$, vetting strategy $VS$, 
   system scores $S=\{s_i\}$, estimator $p_{est}(z)$
   \WHILE{$T\geq 0$}
	   \STATE {Select a batch $B \subseteq U$ according to vetting strategy $VS$.}
	   \STATE {Query oracle to vet $B$ and obtain true annotations $\tilde{z}$.}
	   \STATE {$U = U \setminus B$, $V = V \cup B$}
	   \STATE {$T = T - |B|$}
     \STATE {Fit estimator $p_{est}$ using $U,V,S$.}
   \ENDWHILE
	 \STATE {Estimate performance using $p_{est}(z)$}
\end{algorithmic}
}
\end{algorithm}

\begin{figure}[t!]
    \centering
    \includegraphics[width=0.95\linewidth]{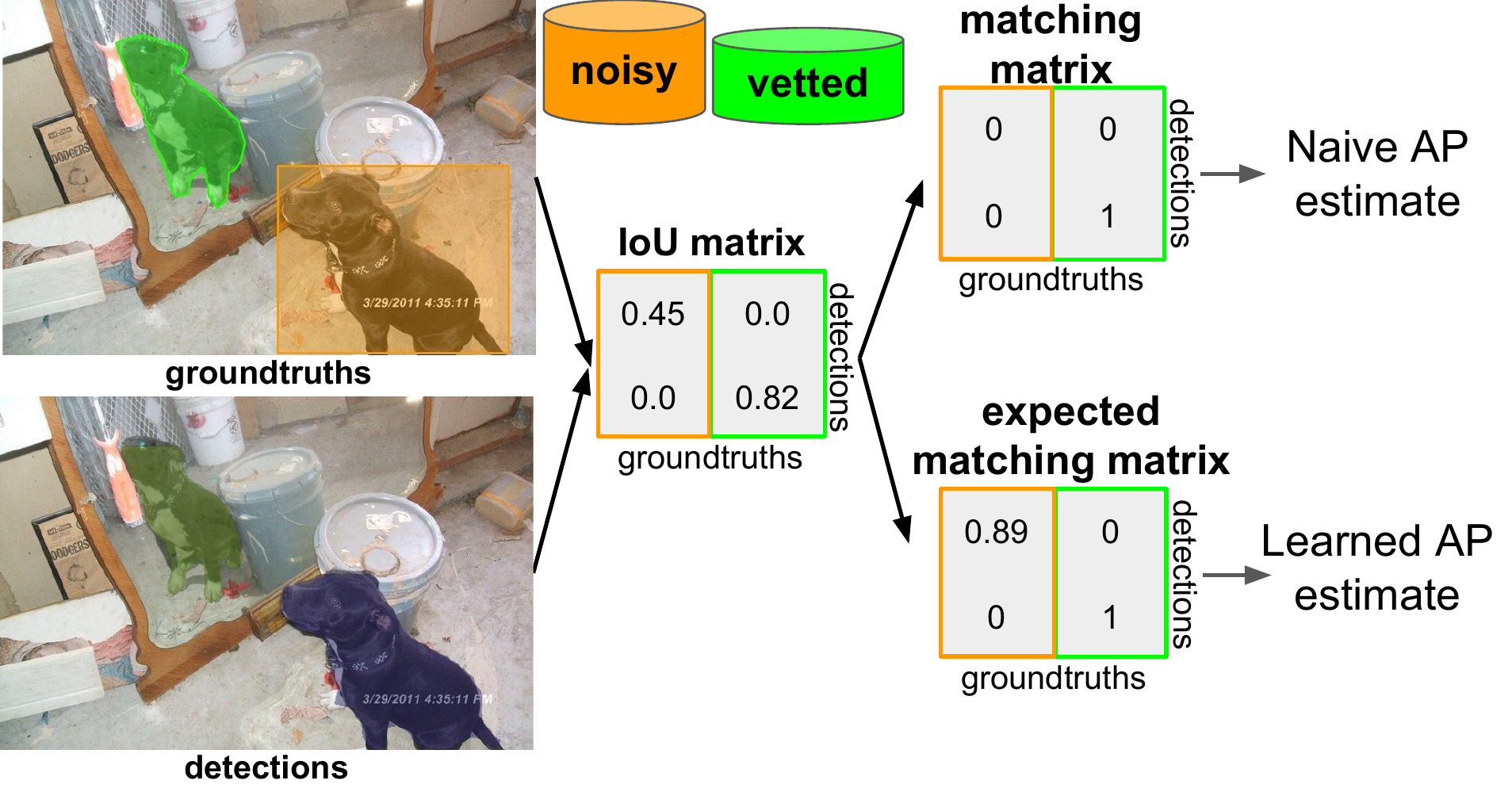}
    \vspace{-0.1in}
    \caption{Standard instance segmentation benchmarks ignore unvetted data
    (top pathway) when computing Average Precision. Our proposed estimator
    for this task computes an expected probability of a match for coarse bounding box annotations when vetted instance masks aren't available.}
\label{fig:estimating_map}
\vspace{-0.1in}
\end{figure}

\begin{figure*}[t!]
    \centering
    \includegraphics[width=0.95\linewidth]{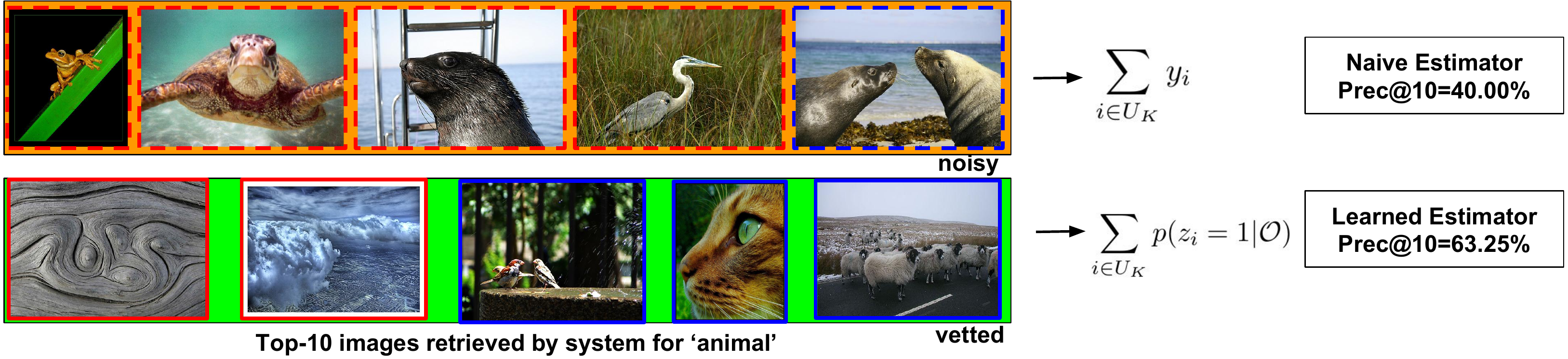}
    \caption{Estimating $Prec@K$. Images at left are the top K=10 entries returned by the system being evaluated. The image border denotes the current label and vetting status (solid blue/red = vetted positive/negative, and dotted blue/red = noisy positive/negative). Estimates of precision can be significantly improved by using a learned estimator trained on the statistics of examples that have already been vetted. Current approaches that evaluate on vetted-only or vetted+noisy labels (naive) produce poor estimates of precision (30\% and 40\%  respectively). Our learned estimator is much closer to the true precision (63\% vs 80\% respectively).}
\label{fig:estimating_prek}
\vspace{-0.1in}
\end{figure*}

Our proposed active testing framework (see Alg 1) estimates this distribution
$p_{est}(z)$ based on available observations and predicts expected benchmark
performance under this distribution:
\begin{align}
  E_{p_{est}(z)} \Big[ Q(z) \Big] \qquad \text{{\bf [Learned Estimator]}} \label{eq:exp}
\end{align}

{\bf Computing expected performance:}
Given posterior estimates $p(z_i | \mathcal{O})$ we can always compute the
expected performance metric $Q$ by generating samples from these distributions,
computing the metric for each joint sample, and average over samples. Here we
introduce two applications (studied in our experiments) where the metric is 
linear or quadratic in $z$, allowing us to compute the expected performance in
closed-form.

{\bf Multi-label Tags:} Multi-label tag prediction is a common task in
video/image retrieval. Following recent work~\cite{joulin2016learning,
gong2013deep,izadinia2015deep, guillaumin2009tagprop}, we measure accuracy
with Precision@K - e.g., what fraction of the top $K$ search results contain
the tag of interest? In this setting, noisy labels $y_i$ come from user 
provided tags which may contain errors and are typically incomplete.
Conveniently, we can write expected performance Eq. \ref{eq:exp} for
Precision@K for a single tag in closed form:
\begin{align} 
E[Prec@K] = \frac{1}{K} \Big( \sum_{i \in V_K} {\tilde z}_{i} + \sum_{i \in U_K} p(z_{i} = 1|\mathcal{O}) \Big) 
\label{eq:prec}
\end{align}
\noindent where we write $V_K$ and $U_K$ to denote the vetted and unvetted
subsets of $K$ highest-scoring examples in the total set $V \cup U$. Some
benchmarks compute an aggregate mean precision over all tags under
consideration, but since this average is linear, one again obtains a closed
form estimate.

{\bf Instance segmentation:} Instance segmentation is another natural task for
which to apply active testing. It is well known that human annotation is
prohibitively expensive -- ~\cite{cordts2016cityscapes} reports that an average
of more than 1.5 {\em hours} is required to annotate a single image. Widely used
benchmarks such as~\cite{cordts2016cityscapes} release small fraction of images annotated with high quality, along with a larger set of noisy or
``coarse''-quality annotations. Other instance segmentation
datasets such as COCO~\cite{lin2014microsoft} are constructed stage-wise by first creating a
detection dataset which only indicates rectangular bounding boxes around each
object which are subsequently refined into a precise instance segmentations.
Fig.~\ref{fig:estimating_map} shows an example of a partially vetted image 
in which some instances are only indicated by a bounding box (noisy), while
others have a detailed mask (vetted).

When computing Average Precision, a predicted instance segmentation is considered a true positive if it has sufficient intersection-over-union (IoU) overlap with a ground-truth instance.  In this setting, we let the
variable $z_i$ indicate that predicted instance $i$ is matched to a
ground-truth instance and has an above threshold overlap. Assuming independence
of $z_i$'s, the expected $AP$ can be written as (see supplement for proof):
\begin{align}
E[AP]&=\frac{1}{N_{p}} \Big(\sum_{k \in V} {\tilde z_k} E[Prec@k] \nonumber \\
     & + \sum_{k \in U} p(z_{k}=1|\mathcal{O}) E[Prec@k] \Big) \label{eq:estimating_average_precision} 
\end{align}

In practice, standard instance segmentation benchmarks are somewhat more complicated. In particular, they enforce one-to-one matching between detections and ground-truth. For example, if two detections overlap a ground-truth instance, only one is counted as a true positive while the other is scored as a false positive. This also holds for multi-class detections - if a detection is labeled as a dog (by matching to a ground-truth dog), it can no longer be labeled as cat.  While this interdependence can in principle be modeled by the conditioning variables $\mathcal{O}$ which could include information about which class detections overlap, in practice our estimators for $p(z_i=1|\mathcal{O})$ do not take this into account. Nevertheless, we show that such estimators provide remarkably good estimates of performance.

{\bf Fitting estimators to partially vetted data:} We alternate between vetting
small batches of data and refitting the estimator to the vetted set. For
multi-label tagging, we update estimates for the prior probability that a noisy
tag for a particular category will be flipped when vetted $p({\tilde z}_i \neq
y_i)$.  For instance segmentation, we train a per-category classifier that
uses sizes of the predicted and unvetted ground-truth bounding box to predict
whether a detected instance will overlap the ground-truth.  We discuss the
specifics of fitting these particular estimators in the experimental results.


\subsection{Vetting Strategies} 

The second component of the active testing system is a strategy for choosing the ``next'' data samples to vet.  The goal of such a
strategy is to produce accurate estimates of benchmark performance with fewest number of vettings.  An alternate, but closely related
goal, is to determine the benchmark rankings of a set of recognition systems
being compared. The success of a given strategy depends on the distribution of
the data, the chosen estimator, and the system(s) under test. We consider
several selection strategies, motivated by existing data collection practice
and modeled after active learning, which adapt to these statistics in order to
improve efficiency. 

\textbf{Random Sampling:} The simplest vetting strategy is to choose test
examples to vet at random. The distribution of examples across categories often follows a long-tail distribution. To achieve faster uniform convergence of performance estimates across all categories, we use a hierarchical sampling approach in which we first sample a category and then select a sub-batch of test examples to vet from that category. This mirrors the way, e.g. image classification and detection datasets are manually curated to assure a minimum number of examples per category.

\textbf{Most-Confident Mistake (MCM):} This strategy selects unvetted examples for which the system under test reports a high-confidence detection/classification score, but which are considered a mistake according to the current metric estimator. Specifically, we focus on the strategy of selecting {\em Most-confident Negative} which is applicable to image/video tagging where the set of user-provided tags are often incomplete.  The intuition is that, if a high-performance system believes that the current sample is a positive with high probability, it's likely that the noisy label is at fault.  This strategy is motivated by experience with object detection benchmarks where, e.g.,  visualizing high-confident false positive face detections often reveals missing annotations in the test set~\cite{mathias2014face}.

\begin{figure*}[t!]
\centering
	\begin{subfigure}[b]{0.49\linewidth}
	\centering
    \includegraphics[width=0.999\linewidth]{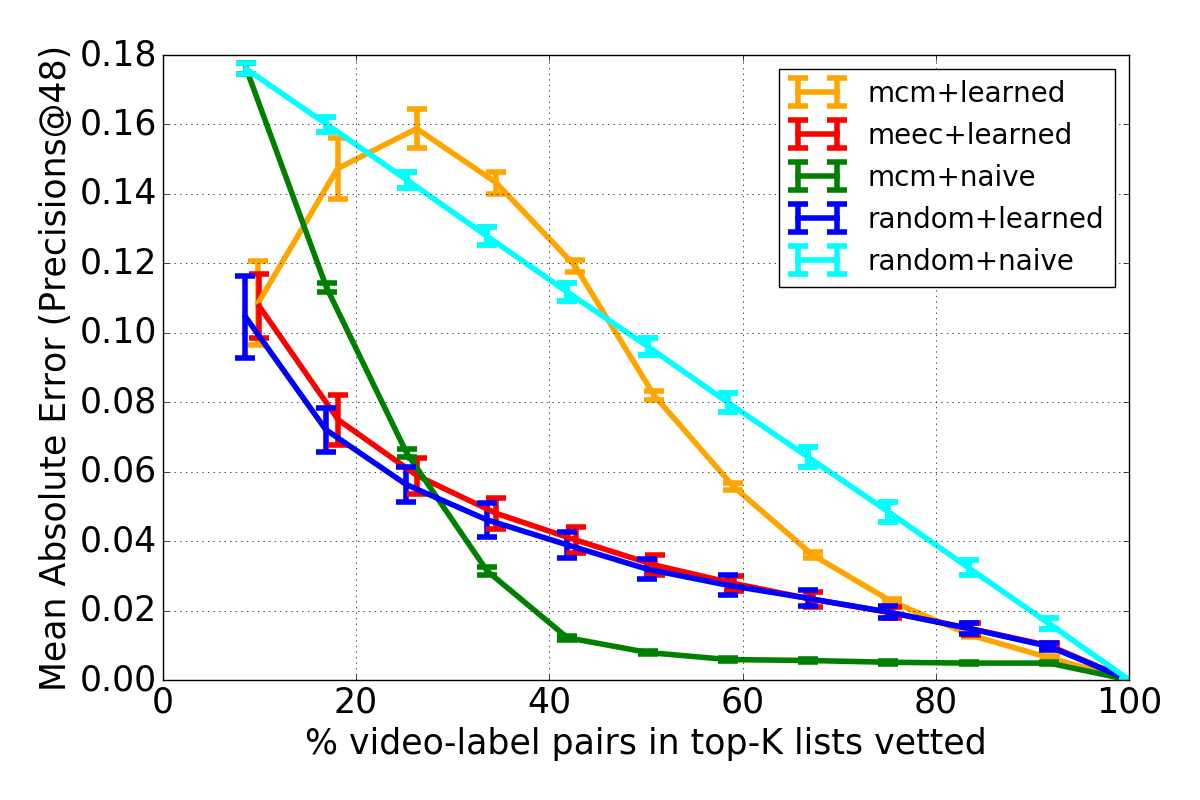}
    \vspace{-0.3in}
	\caption{Micro-Videos}
    \end{subfigure}
    \begin{subfigure}[b]{0.49\linewidth}
    \centering
	\includegraphics[width=0.999\linewidth]{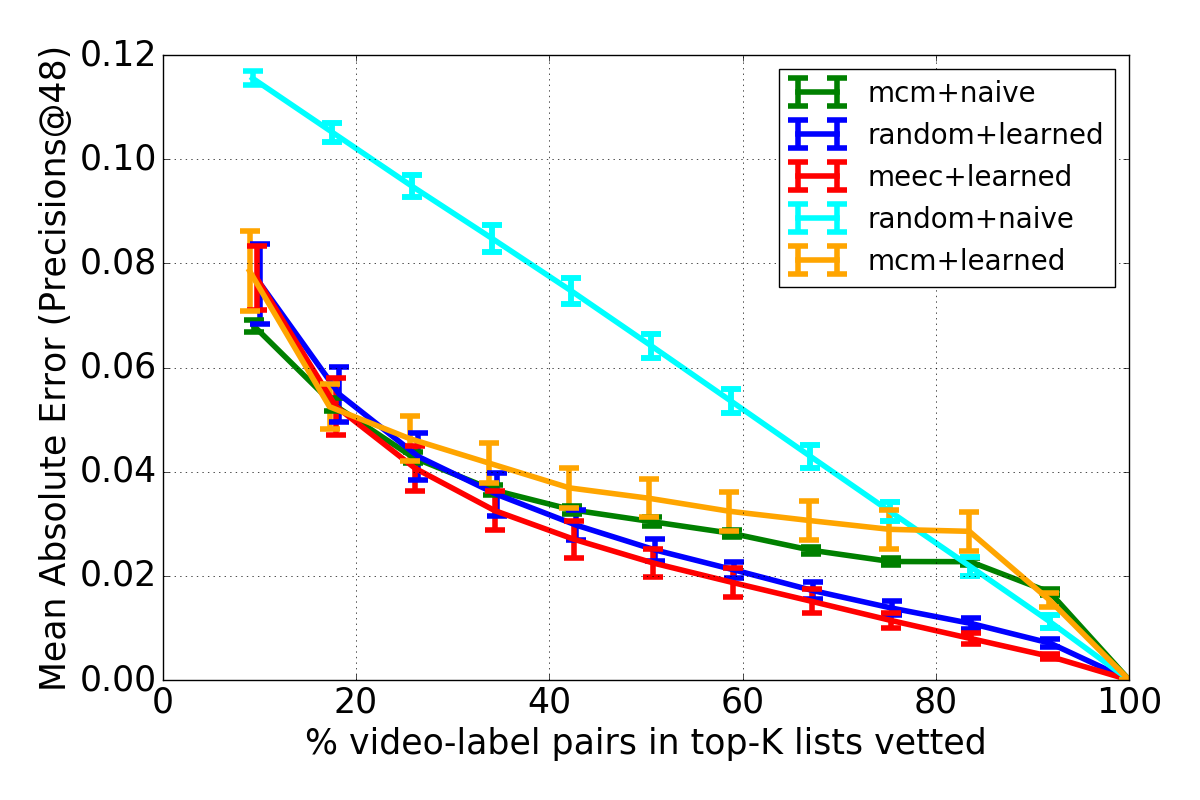}
	\vspace{-0.3in}
    \caption{NUS-WIDE}
    \end{subfigure}
    \vspace{-0.15in}
  \caption{Results for multi-label classification task. The figures show the
  mean and standard deviation of the estimated Precision@48 at different amount
  of annotation efforts. Using a fairly simple estimator and vetting strategy,
  the proposed framework can estimate the performance very closely to the true
  values. For references, the precision@48 averaged across classes is 20.06\%
  and 19.88\% for Microvideos and NUS-WIDE respectively.}
\label{fig:multilabel_classification_simulation_results}
\vspace{-0.2in}
\end{figure*}

\textbf{Maximum Expected Estimator Change (MEEC):} In addition to utilizing the confidence scores produced by the system under test, it is natural to also consider the uncertainty in the learned estimator
$p_{est}(z)$.  Exploiting the analogy of active testing with active learning,
it is natural to vet samples that are most confusing to the current estimator
(e.g., with largest entropy), or ones that will likely generate a large update
to the estimator (e.g., largest information gain). 

Specifically, we explore a active selection strategy based on {\em maximum
expected model change}~\cite{settles2010active}, which in our case corresponds
to selecting a sample that yields the largest expected change in our estimate
of $Q$.  Let $E_{p(z|V)}[Q(z)]$ be the expected performance based on the
distribution $p(z|V)$ estimated from the current vetted set $V$.  $E_{p(z|V
,z_i)}[Q(z)]$ be the expected performance after vetting example $i$ and
updating the estimator based on the outcome.  The {\em actual} change in
the estimate of $Q$ depends on the realization of the random variable $z_i$:

\begin{align}
\Delta_i(z_i) = \Big|E_{p(z|V,z_i)}[Q(z)] - E_{p(z|V)}[Q(z)]\Big|
\end{align}

We can choose the example $i$ with the largest {\em expected} change, using the
current estimate of the distribution over $z_i \sim p(z_i|V)$ to compute the
expected change $E_{p(z_i|V)} \left[\Delta_i(z_i) \right]$.

For $Prec@K$, this expected change is given by:
\vspace{-0.05in}
\begin{eqnarray}
E_{p(z_i|V)}\left[\Delta_i (z_i)\right] = \frac{2}{K} p_i (1-p_i)
\end{eqnarray}
\noindent where we write $p_i = p(z_i = 1| \mathcal{O})$. Interestingly,
selecting the sample yielded the maximum expected change in the estimator
corresponds to a standard {\em maximum entropy} selection criteria for active
learning.  Similarly, in the supplement we show that for $AP$:

\vspace{-0.25in}
\begin{eqnarray}
E_{p(z_i|V)}\left[\Delta_i (z_i)\right] = \frac{1}{N_p} r_i p_i(1-p_i)
\end{eqnarray}
\vspace{-0.2in}

where $r_i$ is the proportion of unvetted examples scoring higher than example
$i$. In this case, we select an example to vet which has high-entropy and for which there is a relatively small proportion of higher-scoring unvetted examples.



\vspace{-0.075in}
\section{Experiments}
\label{sec:exp}
We validate our active testing framework on two specific applications,
multi-label classification and instance segmentation.  For each of these applications, we describe the datasets and systems evaluated and the specifics of the estimators and vetting strategies used.


\vspace{-0.075in}
\subsection{Active Testing for Multi-label Classification}
\noindent
\textbf{NUS-WIDE:} This dataset contains 269,648 Flickr images with 5018 unique
tags. The authors also provide a 'semi-complete' ground-truth via manual
annotations for 81 concepts. We removed images that are no longer available and
images that doesn't contain one of the 81 tags. We are left with around 100K
images spanning across 81 concepts. ~\cite{izadinia2015deep} analyzed the noisy
and missing label statistics for this dataset. Given that the tag is relevant
to the image, there is only 38\% chance that it will appear in the noisy tag
list.  If the tag does not apply, there's 1\% chance that it appears anyway.
They posited that the missing tags are either non-entry level categories (e.g.,
person) or they are not important in the scene (e.g., clouds and buildings).

\textbf{Micro-videos:} Micro-videos have recently become a prevalent form of
media on many social platforms, such as Vine, Instagram, and Snapchat.
~\cite{nguyen2016open} formulated a multi-label video-retrieval/annotation task
for a large collection of Vine videos. They introduce a micro-video dataset,
\textbf{MV-85k} containing 260K videos with 58K tags. This dataset, however,
only provides exhaustive vetting for a small subset of tags on a small subset
of videos. We vetted ~26K video-tag pairs from this dataset, spanning 17503
videos and 875 tags. Since tags provided by users have little constraints, this
dataset suffers from both under-tagging and over-tagging.  Under-tagging comes
from not-yet popular concepts, while over-tagging comes from the spamming of extra  tags In our experiments we use a subset of 75 tags.

\textbf{Recognition systems:} To obtain the classification results, we implement
two multi-label classification algorithms for images (NUSWIDE) and videos
(Microvideos). For NUS-WIDE, we trained a multi-label logistic regression model
built on the pretrained ResNet-50~\cite{he2016deep} features. For Micro-videos,
we follow the state-of-the-art video action recognition
framework~\cite{WangXWQLTV16} modified for the multi-label setting to use 
multiple logistic cross-entropy losses.

\textbf{Learned Estimators:} We use Precision@48 as a evaluation metric.  For
tagging, we estimate the posterior over unvetted tags, $p(z_i | \mathcal{O})$,
based on two pieces of observed information: the statistics of noisy labels
$y_i$ on vetted examples, and the system confidence score, $s_{i}$.  This
posterior probability can be derived as (see supplement for proof):
\begin{align}
	p(z_{i}|s_{i},y_{i}) &=
    \frac{p(y_{i}|z_{i})p(z_{i}|s_{i})}{\sum_{v\in\{0,1\}}p(y_{i}|z_{i}=v)p(z_{i}=v|s_{i})}
\end{align}
Given some vetted data, we fit the tag-flipping priors $p(y_{i}|z_{i})$ by
standard maximum likelihood estimation (counting frequencies). The posterior
probabilities of the true label given the classifier confidence score,
$p(z_{i}|s_{i})$, is fit using logistic regression.

\vspace{-0.1in}
\subsection{Object Instance Detection and Segmentation}

\textbf{COCO Minival:} For instance segmentation, we use `minival2014' subset of
the COCO dataset~\cite{lin2014microsoft}. This subset contains 5k images
spanning over 80 categories. We report the standard COCO metric: Average
Precision (averaged over all IoU thresholds).

To systematically analyze the impact of evaluation on noise and vetting, we
focus evaluation efforts on the high quality test set, but simulate noisy
annotations by replacing actual instance segmentation masks by their
tight-fitting bounding box (the unvetted ``noisy'' set). We then simulate
active testing where certain instances are vetted, meaning the bounding-box is
replaced by the true segmentation mask.

\textbf{Detection Systems:} We did not implement instance segmentation
algorithms ourselves, but instead utilized three sets of detection mask results
produced by the authors of Mask R-CNN~\cite{he2017mask}. These were produced by
variants of the instance segmentation systems proposed
in~\cite{xie2017aggregated,lin2017feature,he2017mask}.

\begin{figure}[t]
	\centering
	\includegraphics[height=0.65\linewidth]{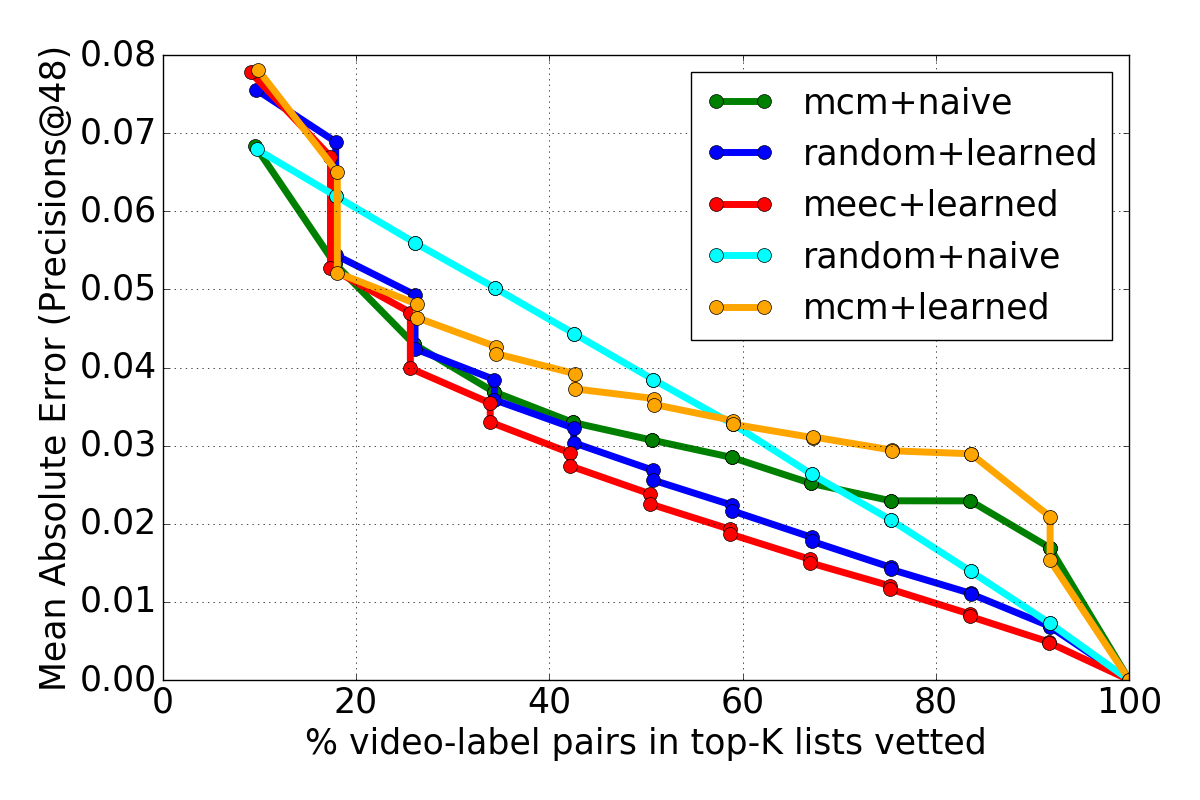}
	\vspace{-0.2in}
    \caption{Decoupling the effect of model change and vetting effort for
    NUS-WIDE. This figure shows the reduction in estimation errors. The
    vertical drop at the same \% vetted point indicates the reduction due to
    estimator quality. The slope between adjacent points indicates value of
    vetting examples. A steeper slope means the strategy is able to obtain a
    better set. In some sense, traditional active learning is concerned primarily with the vertical drop (i.e. a better model/predictor), while active testing also takes direct advantage of the slope (i.e. more vetted labels).}
\label{fig:decouple}
\vspace{-0.15in}
\end{figure}

\begin{figure}[t!]
  \centering
  \includegraphics[width=0.99\linewidth]{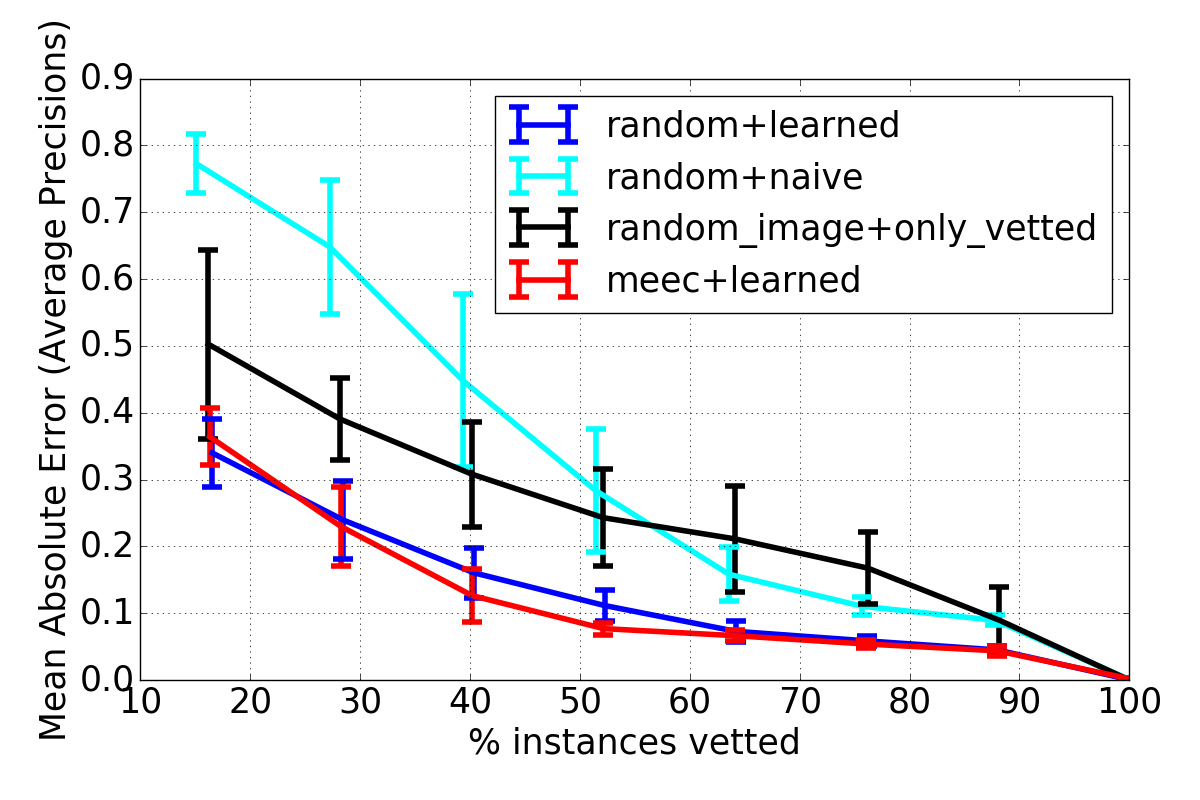}
  \vspace{-0.2in}
  \caption{Results for instance segmentation. With $50\%$ of
  instances vetted, our best model's estimation is $1\%$ AP off from the true
  values with the standard deviation $\le 1\%$. A smart estimator with a
  smarter querying strategy can make the approach more robust and efficient.
  Our approach has better approximation and is less prone to sample bias
  compared to the standard approach("random image"+ "only vetted").}
\label{fig:instance_segmentation_progression}
\vspace{-0.15in}
\end{figure}

\begin{figure*}[t!]
    \centering
    \begin{subfigure}[b]{0.49\linewidth}
    \centering
		\includegraphics[width=0.98\linewidth]{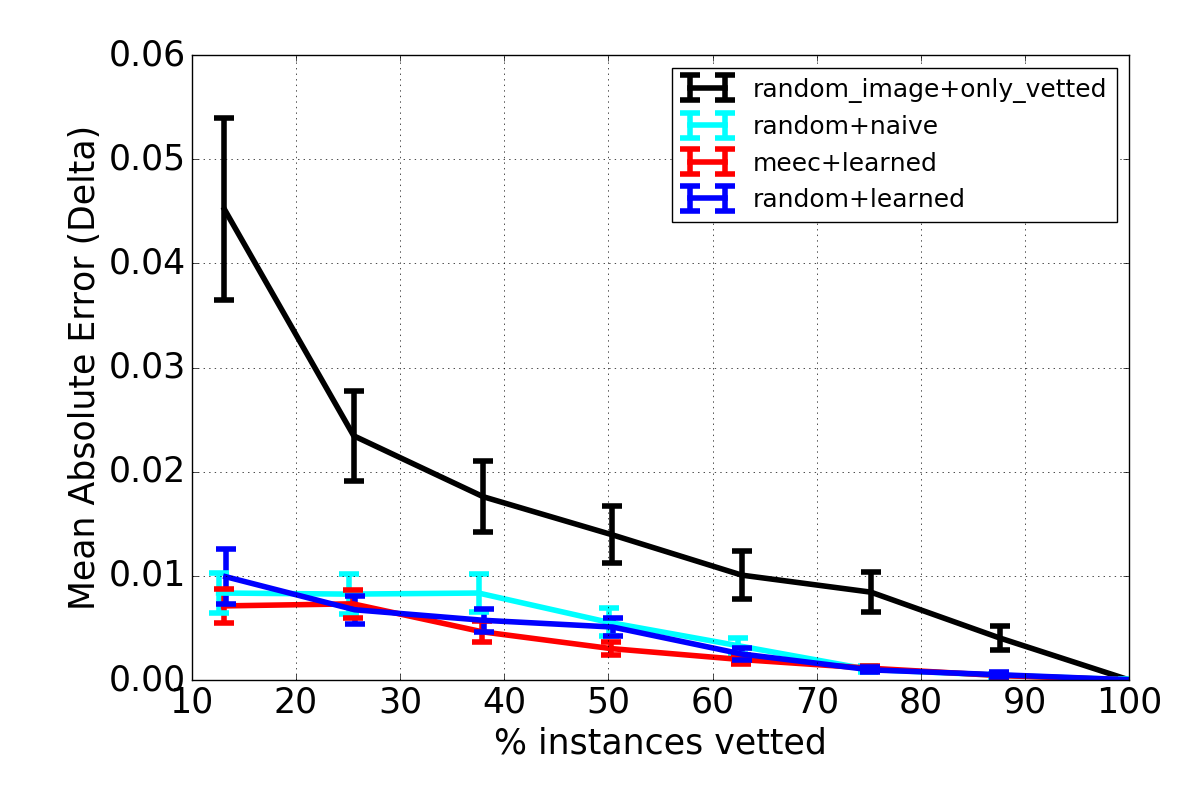}
        \label{fig:multi_model_delta_distance}
    \end{subfigure}
    \begin{subfigure}[b]{0.49\linewidth}
    \centering
    	\includegraphics[width=0.98\linewidth]{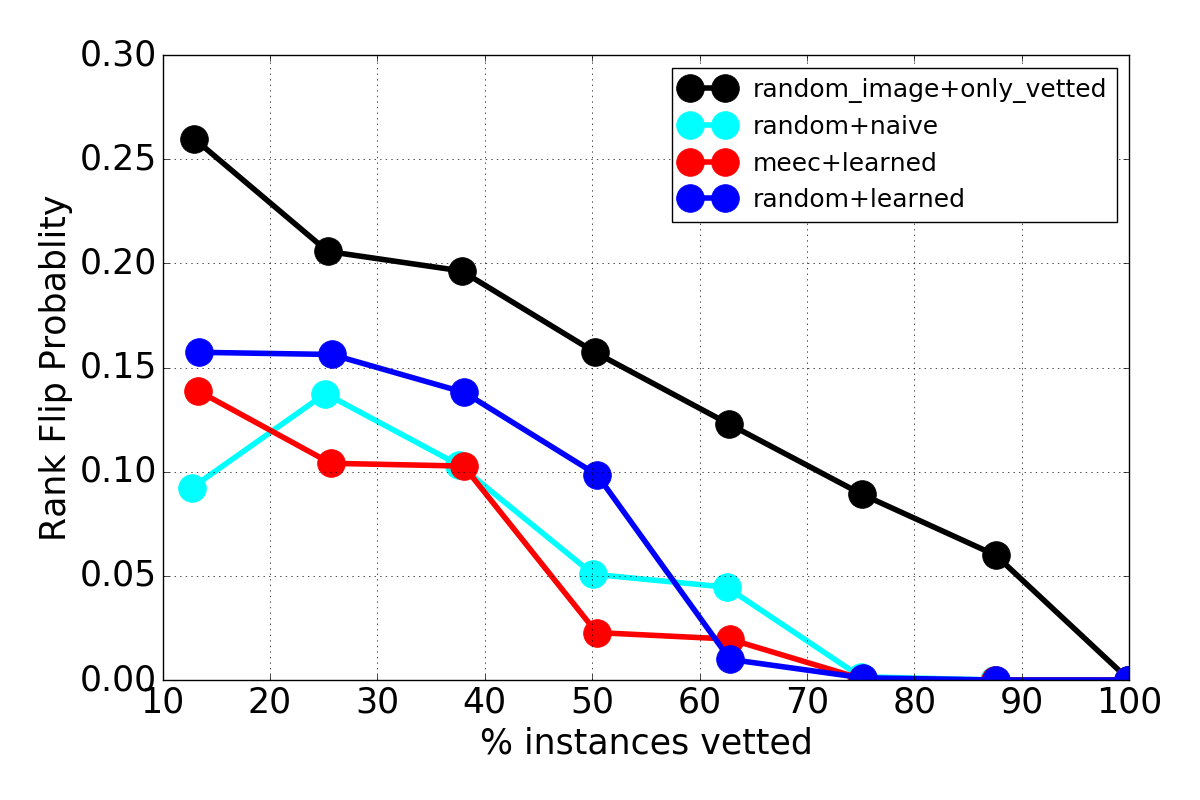}
    	\label{fig:multi_model_prob_flip}
    \end{subfigure}
    \vspace{-0.1in}
    \caption{Relative performance differences and their relative ranking for multiple input systems. The left plot shows the mean squared errors between the current difference to the true difference. The right plot shows how often the ranking orders between two input algorithms are flipped. Both figures suggest that our active testing framework is a more robust and
    efficient approach toward comparing models. With 50\% of the data vetted, standard approaches that evaluate on only vetted data (black curve) incorrectly rank algorithms 16\% of the time, while our learned estimators with active vetting (red curve) reduce this error to 3\% of the time.
    \label{fig:multi_model_instance_segmentation}
    }
\label{fig:multimodel_analyses}
\vspace{-0.15in}
\end{figure*}

\textbf{Learned Estimators:} To compute the probability whether a detection will
pass the IoU threshold with a bounding box unvetted ground-truth instance
($p(z_i | \mathcal{O})$ in Eq.~\ref{eq:estimating_average_precision}), we train
a $\chi^2$-SVM using the vetted portion of the database. The features for an
example includes the category id, the `noisy' IoU estimate, the size of the
bounding box containing the detection mask and the size of ground-truth bounding
box. The training label is true whether the true IoU estimate, computed using
the vetted ground-truth mask and the detection masks, is above a certain input
IoU threshold.


\vspace{-0.10in}
\subsection{Efficiency of active testing estimates} 

We measure the estimation accuracy of different combination of vetting
strategies and estimators at different amount of vetting efforts. We compute
the absolute error between the estimated metric and the true (fully vetted)
metric and average over all classes. Averaging the absolute estimation error
across classes prevents over-estimation for one class canceling out
under-estimation from another class.  We plot the mean and the standard
deviation over 50 simulation runs of each active testing approach.

\textbf{Performance estimation:}
Figure~\ref{fig:multilabel_classification_simulation_results} shows the results
for estimating $Prec@48$ for NUSWIDE and Microvideos. The x-axis indicates the
percentage of the top-k lists that are vetted. For the $Prec@K$ metric, it is
only necessary to vet 100\% of the top-k lists rather than 100\% of the whole
test set\footnote{The ``vetted only'' estimator is not
applicable in this domain until at least $K$ examples in each short list
have been vetted and hence doesn't appear in the plots.}.  A 'random' strategy with a `naive' estimator follows a linear trend since each batch of vetted examples contributes on average the same reduction in estimation error.  The most confident mistake (mcm) heuristic works very well for Microvideos due to the substantial amount of under-tagging.  However, in more reasonable balanced settings such as NUS-WIDE, this heuristic does not perform as well. The MCM vetting strategy does not pair well with a learned estimator due to its biased sampling which quickly results in priors that overestimate the number missing tags.  In contrast, the random and active {\it MEEC  }vetting strategies offer good samples for learning a good estimator. At 50\% vetting effort, {\it MEEC } sampling  with a learned estimator on average can achieve within 2-3\% of the real estimates.

Figure~\ref{fig:decouple} highlights the relative value of establishing the
true vetted label versus the value of vetted data in updating the estimator. In some sense, traditional active learning is concerned primarily with the vertical drop (i.e. a better model/estimator), while active testing also takes direct advantage of the slope (i.e. more vetted labels).
The initial learned estimates have larger error due to small sample size, but
the fitting during the first few vetting batches rapidly improves the estimator
quality. Past 40\% vetting effort, the estimator model parameters stabilize and
remaining vetting serves to correct labels whose true value can't be predicted
given the low-complexity of the estimator.

Figure~\ref{fig:instance_segmentation_progression} shows similar results for
estimating the mAP for instance segmentation on COCO. The current `gold
standard' approach of estimating performance based only on the vetted subset of
images leads to large errors in estimation accuracy and high variance from from
small sample sizes.  In the active testing framework, input algorithms are
tested using the whole dataset (vetted and unvetted). Naive estimation is
noticeably more accurate than vetted only and the learned estimator with
uncertainty sampling further reduces both the absolute error and the variance.

\textbf{Model ranking:}
The benefits of active testing are highlighted further when we consider the
problem of ranking system performance.  We are often interested not in the
absolute performance number, but rather in the performance gap between
different systems. We find that active testing is also valuable in this
setting.  Figure~\ref{fig:multi_model_instance_segmentation} shows the error in
estimating the {\em performance gap} between two different instance
segmentation systems as a function of the amount data vetted. This follows a
similar trend as the single model performance estimation plot. Importantly, it
highlights that only evaluating vetted data, though unbiased, typically
produces a large error in in performance gap between models to high variance  
in the estimate of each individual models performance. In particular, if we
use these estimates to rank two models, we will often make errors in model 
ranking even when relatively large amounts of the data have been vetted.
Using stronger estimators, actively guided by {\it MEEC} sampling provide
accurate rankings with substantially less vetting effort. With 50\% of the data vetted, standard approaches that evaluate on only vetted data (black curve) incorrectly rank algorithms 15\% of the time, while our learned estimators with active vetting (red curve) reduce this error to 3\% of the time.

\textbf{Conclusions} We have introduced a general framework for active testing
that minimizes human vetting effort by actively selecting test examples to
label and using performance estimators that adapt to the statistics of the test
data and the systems under test. Simple implementations of this concept
demonstrate the potential for radically decreasing the human labeling effort
needed to evaluate system performance for standard computer vision tasks.
We anticipate this will have substantial practical value in the ongoing 
construction of such benchmarks.

\textbf{Acknowledgement} We thank Piotr Dollar and Ross Girshick for providing the instance segmentation results for the COCO dataset. This project was supported in part by NSF grants IIS-1618806 and IIS-1253538.

\bibliography{activetesting}
\bibliographystyle{icml2018}

\end{document}